\pdfoutput=1
\documentclass[11pt]{article}

\usepackage[final]{acl}

\usepackage{times}
\usepackage{latexsym}

\usepackage[T1]{fontenc}
\usepackage[utf8]{inputenc}
\usepackage{microtype}
\usepackage{tcolorbox,paralist,multirow}
\usepackage{array}
\usepackage{booktabs} 
\usepackage{longtable}
\usepackage{supertabular}

\usepackage{microtype}

\usepackage{inconsolata}

\usepackage{graphicx}

\title{Exploring Personality-Aware Interactions in Salesperson Dialogue Agents}

\author{Sijia Cheng\quad Wen-Yu Chang\quad Yun-Nung Chen\\
  National Taiwan University, Taipei, Taiwan\\
  \texttt{\{r11922184, f10946031\}@csie.ntu.edu.tw\quad y.v.chen@ieee.org}\\}

\begin{document}
\maketitle
\begin{abstract}
The integration of dialogue agents into the sales domain requires a deep understanding of how these systems interact with users possessing diverse personas. 
This study explores the influence of user personas, defined using the Myers-Briggs Type Indicator (MBTI), on the interaction quality and performance of sales-oriented dialogue agents. 
Through large-scale testing and analysis, we assess the pre-trained agent's effectiveness, adaptability, and personalization capabilities across a wide range of MBTI-defined user types. 
Our findings reveal significant patterns in interaction dynamics, task completion rates, and dialogue naturalness, underscoring the future potential for dialogue agents to refine their strategies to better align with varying personality traits.
This work not only provides actionable insights for building more adaptive and user-centric conversational systems in the sales domain but also contributes broadly to the field by releasing persona-defined user simulators. 
These simulators, unconstrained by domain, offer valuable tools for future research and demonstrate the potential for scaling personalized dialogue systems across diverse applications.\footnote{The code and scripts are available at \url{https://github.com/MiuLab/MBTI-User}.}

\end{abstract}

\section{Introduction}

Dialogue systems are becoming essential in the sales industry, enabling businesses to communicate more effectively with their customers \cite{adamopoulou2020overview}. 
These AI-driven systems assist customers by providing product recommendations, answering inquiries, and supporting better purchasing decisions. 
However, each user is unique, with communication styles that vary according to their personality traits. 
To create more effective dialogue systems, it is crucial to account for these individual differences.
Most conventional conversational systems adopt a ``one size fits all'' approach, which often results in impersonal interactions. 
Incorporating an understanding of personality differences can enable dialogue systems to tailor their responses and communication styles, enhancing user satisfaction and engagement.

Personality significantly influences conversational preferences. 
While some individuals favor concise, direct exchanges, others value detailed and exploratory interactions. 
For instance, \citet{zhang2018personalizingdialogueagentsi} proposed creating user profiles based on 5-sentence textual descriptions, such as:
``I am a vegetarian. I like swimming. My father used to work for Ford. My favorite 
band is Maroon5. I got a new job last month, which is about advertising design.''
Current dialogue systems often rely on general user information, such as name, age, preferences, and occupation, to personalize interactions. 
However, these approaches failed to capture the deeper psychological traits that drive user behavior and shape user preference.

To fill in the gap, this study investigates how personality traits, as defined by the Myers-Briggs Type Indicator (MBTI), influence interaction quality and the performance of sales-oriented dialogue agents.
The MBTI is a widely used framework that categorizes individuals into 16 personality types based on four key dimensions: extraversion (E) vs. introversion (I), sensing (S) vs. intuition (N), thinking (T) vs. feeling (F), and judging (J) vs. perceiving (P). 
These dimensions shape how individuals gather information, make decisions, and respond to various communication styles.

In this study, we employed \textsc{SalesAgent} \cite{chang-chen-2024-injecting}, a pretrained sales-oriented dialogue agent, to investigate how different personality traits influence conversational outcomes. 
The goal was to explore whether variations in user personalities lead to distinct interaction patterns, thereby highlighting the potential for developing customized dialogue agents tailored for marketing applications~\cite{chiu-etal-2022-salesbot,murakhovska-etal-2023-salespeople}.
\textsc{SalesAgent}~\cite{chang-chen-2024-injecting} is designed to initiate interactions with casual chit-chat before transitioning into task-oriented conversations.
We conducted experiments using \textsc{SalesAgent} with a diverse set of MBTI-based user simulators. 
Each simulator represented a specific personality type, characterized by hobbies, occupations, and preferences reflective of their MBTI profile. 
The agent's marketing performance was evaluated across key metrics, including task completion rates and dialogue quality (measured by naturalness, coherence, smoothness, agent aggressiveness, and consistency). 
Our analysis uncovered significant differences in how users with various personality traits interacted with the agent.
For instance, some personality types responded positively to direct, structured communication, while others preferred more flexible, open-ended interactions.

Our findings highlight the importance of personality-based approaches in enhancing dialogue agents. 
The results revealed clear distinctions in user-agent interactions based on MBTI personality types. 
Extraverted (E) users achieved the highest task success rate at 82.7\%, while judging (J) users had the lowest at 62.1\%. 
Similarly, feeling (F) users sustained conversations on the same topic longer, with a continuation ratio of 40.58\%, compared to 31.30\% for judging (J) users. 
These findings demonstrate the need for dialogue agents to adapt their strategies to accommodate diverse personality traits effectively.

Beyond providing practical recommendations, this study contributes to the broader field of personalized conversational AI by introducing a set of MBTI-based persona-defined user simulators. These simulators serve as a valuable resource for future research, enabling the development of more personalized systems across various domains, such as psychological counseling. The insights gained from this work demonstrate the potential for expanding personalized dialogue systems to a wide range of applications.

\section{Related Work}
\subsection{LLMs in Conversational Sales}
Previous studies have been explored various perspectives in terms of application in business domain, such as marketing, sales etc. \citet{chiu-etal-2022-salesbot} developed the first salesbot datasets combining chit-chat dialogue and task-oriented dialogue to mimick the conversation between actual salesperson and the user. \citet{chang-chen-2024-injecting} further improved the datasets, SalesBot 2.0, by leveraging LLMs common-sense and further construct a \textsc{SalesAgent} by fine-tuning LLaMA-2-7B model with the new datasets. Murakhovs’ka et al.~\cite{murakhovska-etal-2023-salespeople} proposed SalesOps, a framework that uses LLM-powered agents to simulate realistic sales conversations involving complex products. Their system, SalesBot, integrates product catalogs and buying guides to provide both recommendations and educational value. Compared to professional salespeople, SalesBot showed similar fluency and informativeness but still underperformed in recommendation accuracy. This work highlights the potential of LLMs in personalized marketing and the importance of knowledge grounding in sales-oriented dialog systems.

\subsection{Personality Research in LLMs}

Recent studies have categorized the use of personas in large language models (LLMs) into two primary directions: LLM Role-Playing and LLM Personalization~\cite{tseng2024talespersonallmssurvey}. In the role-playing setting, personas are assigned to LLMs to simulate specific professional or social roles (e.g., judges, doctors, engineers) within task-oriented environments. In contrast, personalization focuses on modeling user personas to generate tailored responses in applications such as recommendation systems, dialogue agents, and educational platforms. This dichotomy offers a unified framework for understanding how persona modeling enhances contextual relevance and user interaction in LLM-driven systems.

In the domain of role-playing,  \citet{park2023generativeagentsinteractivesimulacra} introduced generative agents—LLM-based agents equipped with memory, reflection, and planning capabilities—to emulate human-like personas in interactive environments. Deployed in a sandbox world, these agents exhibited believable individual and social behaviors, including relationship formation and event coordination. Their architecture highlights the importance of dynamic memory and self-reflection in supporting consistent and evolving personas.

For personalization, researchers have explored leveraging personality traits to guide LLM behavior. \citet{du2024reinforcement} proposed the RLLI framework, which employs LLM-based generative agents to simulate user feedback based on the Big Five personality model. These agents generate subjective quality-of-experience (QoE) ratings, enabling reinforcement learning models to adapt AIGC services to user preferences. By embedding personality traits through prompt engineering, the system enables scalable, human-like personalization without requiring real-time human feedback.

Moreover, \citet{jiang2024personallminvestigatingabilitylarge} developed PersonaLLM, a framework for evaluating whether LLMs can consistently express assigned Big Five traits. Through BFI tests and narrative writing, they showed that GPT-3.5 and GPT-4 can exhibit trait-consistent linguistic behaviors, which are partially recognizable by human raters. Interestingly, the perceived personality diminishes when participants are told the author is an AI, highlighting the role of transparency in AI-human interaction.

Finally, \citet{pan2023llmspossesspersonalitymaking} investigated whether LLMs inherently exhibit human-like personalities using the MBTI as an evaluation framework. Their findings suggest that LLMs, particularly GPT-4, can display consistent MBTI types (e.g., INTJ), and that these personalities can be influenced by prompt design and training data. While informal, MBTI thus serves as a potentially useful diagnostic lens for analyzing LLM behavior.

\subsection{Salesperson Dialogue Agent}
\citet{chang-chen-2024-injecting} introduced \textsc{SalesAgent}, a dialogue agent powered by LLMs and specifically designed to employ sales-oriented conversational strategies. The approach fine-tunes the agent using automatically generated internal thoughts aligned with an expert-defined conversational framework, enabling the agent to exhibit strategic and purposeful dialogue behaviors.

This method builds on the chain-of-thought (CoT) and ReAct prompting paradigms~\cite{wei2022chain,yao2023react}. Within this framework, the dialogue agent begins by analyzing the conversational context to generate an understanding of the current dialogue state. This includes recognizing whether the user has revealed specific intents or shown interest in particular topics. Based on this contextual understanding, the agent formulates a dialogue policy that guides the generation of coherent and strategic responses, steering the conversation toward topics likely to engage the user effectively.

\textsc{SalesAgent} was evaluated through simulated conversations with 50 user simulators featuring diverse personas generated by LLMs. These personas included variations in occupation, hobbies, and interest levels across different topics. The results demonstrated that CoT strategies improved the agent’s ability to smoothly transition between topics while maintaining explainability and coherence in its responses.

Despite its promising results, the study's experimental setup did not account for the influence of users' personal traits within the personas of the simulators. While the approach leveraged general attributes such as \emph{occupations}, \emph{hobbies}, or \emph{interest levels} to define user personas, these descriptors are insufficient to fully encapsulate the nuanced and multifaceted nature of individual personalities. 
Personality is a critical factor that shapes how users interact and respond in conversations, and relying solely on generic characteristics fails to capture the diversity and complexity of real user behaviors.

To address this limitation, our work adopts a more structured and comprehensive approach by incorporating the Myers-Briggs Type Indicator (MBTI)~\cite{boyle1995myers} to define user personas. MBTI categorizes personality traits across four dimensions: \textit{extroversion/introversion}, \textit{sensing/intuition}, \textit{thinking/feeling}, and \textit{judging/perceiving}, resulting in 16 unique personality types. By using MBTI-defined personas in our user simulators, we aim to create more realistic and nuanced representations of individual differences.

This study enables us to explore whether adapting dialogue strategies to users with distinct personality profiles and interaction styles is important.
By examining the interplay between user personalities and agent behaviors, we seek to gain deeper insights into the agent's ability to engage effectively with a diverse range of users. 
Furthermore, this allows us to evaluate the agent's performance in handling varied conversational dynamics and to propose improvements for tailoring dialogue systems to better meet individual needs in the future.

\section{Personality-Defined User Simulation}
To investigate how personality influences user behavior, we examine the interactions between \textsc{SalesAgent} and personality-defined user simulators~\cite{li2016user,gur2018user}. 
This section outlines the process of constructing user simulators with assigned personality traits and presents statistics on the resulting user profiles.

\subsection{Myers-Briggs Type Indicator (MBTI)}
The Myers-Briggs Type Indicator (MBTI) is a widely recognized personality assessment tool that categorizes individuals into 16 personality types based on four key dimensions \cite{boyle1995myers}: 
\begin{compactitem} 
\item Extraversion (E) vs. Introversion (I) 
\item Sensing (S) vs. Intuition (N) 
\item Thinking (T) vs. Feeling (F) 
\item Judging (J) vs. Perceiving (P) 
\end{compactitem}
Developed by Katharine Cook Briggs and Isabel Briggs Myers and grounded in Carl Jung's psychological type theory, MBTI has found broad applications in career counseling, organizational behavior, and human-computer interaction \cite{kuipers2009influence, garden1997relationships}.

MBTI is particularly advantageous in computational research due to its structured and categorical nature, allowing personality modeling to be framed as a multi-class classification problem \cite{lacava2024openmodelsclosedminds}. Compared to other personality frameworks, MBTI provides an intuitive and actionable means of analyzing behavioral traits, making it a suitable foundation for AI-driven applications \cite{stajner-yenikent-2021-mbti}.

Our work focuses on analyzing a \emph{single} dimension of MBTI at a time. This approach enables us to isolate and differentiate personality traits within the user simulator and assess the impact of these traits on the performance of sales-oriented agents.

\subsection{MBTI-Defined Personality}
While there have been numerous attempts to leverage MBTI for response generation \cite{fu2024styemp,wu2025traits}, most prior work has focused on chit-chat scenarios rather than goal-oriented settings. 
Our work is the first to design MBTI-defined user simulators specifically for evaluating dialogue agents' performance in communication within task-oriented contexts.

This study emphasizes the incorporation of MBTI personality traits into user simulators to create diverse and realistic interaction scenarios. Unlike traditional approaches that rely on fixed, predefined personality profiles, our method focuses on generating user personas based on \emph{individual} MBTI dimensions: Extraversion (E), Introversion (I), Sensing (S), Intuition (N), Thinking (T), Feeling (F), Judging (J), and Perceiving (P). This approach enables a more flexible and nuanced representation of user behaviors, capturing the variability inherent in different personality traits.

To construct these personas, we use structured prompts that emphasize the defining characteristics of each MBTI dimension. For instance, in generating personas for the Extraverted (E) type, prompts highlight attributes such as a focus on external interactions, gaining energy from social engagement, and readily taking initiative. Key characteristics for the Extraverted (E) type include:
\begin{compactitem}
    \item Focus on the outside world 
    \item Gain energy by interacting with people 
    \item Take action quickly 
    \item Communicate through talking
    \item Process ideas outwardly 
    \item Act before thinking it through 
    \item Readily take initiative 
    \item Have many broad interests 
\end{compactitem}

%The following is the JSON format used:
%\begin{verbatim}
%[
%    {"id": "<id>", "persona": "<persona 1>"},
%    ...
%]
%\end{verbatim}
The personas generated through these prompts offer valuable insights into how individuals with different personality traits navigate social interactions and make decisions. Below, we present the complete prompt used to instruct ChatGPT to generate user profiles and and one example of the generated user profiles for the Extraverted (E) type.

\begin{tcolorbox}[width=\columnwidth,colback=white]
\small
\begin{verbatim}
Create a set of personas for a user
simulator, each embodying the Extraverted 
personality trait. Each should be an 
introduction of his or herself with the 
hobby, job, and characteristics.

People who prefer Extraversion (E) tend to:
Focus on the outside world
Gain energy by interacting with people
Take action quickly
Communicate through talking
Process ideas outwardly
Act before thinking it through
Readily take initiative
Have many broad interests

Objective:
The aim is to create well-rounded personas 
that capture the essence of extraverted 
personality traits. These personas should
provide insight into how extraverted 
individuals engage with others, manage 
social interactions, and thrive in 
environments that allow them to express 
their outgoing nature.

Format:
Please generate personas in JSON format 
like the following: 
[
{ "id": "<id>, 
 "persona": "<persona_1>"}, ... 
] 

Sample output: 
<sample output>
\end{verbatim}
\end{tcolorbox}

To ensure compatibility with our simulator framework, the personas are formatted in a JSON structure. Each persona includes a unique identifier and a brief introduction that outlines hobbies, occupations, and key personality characteristics. For a comprehensive overview, the complete set of personality keywords corresponding to all eight MBTI dimensions is provided in Appendix~\ref{sec:keywords}.

A sample user profile is provided here as a reference. It includes detailed information describing an E-type personality, which can be used to prompt LLMs to role-play and simulate the target user.
\begin{tcolorbox}[width=\columnwidth,colback=white]
\small
\begin{verbatim}
You're Mia Gomez, a 27-year-old event 
planner who thrives on creating memorable 
experiences. Whether coordinating weddings, 
festivals, or corporate functions, you handle
every detail with flair and precision. You 
love meeting new people, exploring art 
galleries, and dancing at local clubs. 
Your vibrant personality and ability to 
connect with anyone make you a favorite 
among clients and friends alike.
\end{verbatim}
\end{tcolorbox}

\subsection{Persona Dataset Statistics}

To create a diverse and representative user simulation, we generated a total of 80 personas, with 10 personas assigned to each of the 8 MBTI dimensions. The dataset reflects a wide range of characteristics, including variations in age and occupational backgrounds.
Note that the age and occupation distributions can be adjusted through sampling; however, we did not predefine a specific distribution for this study, allowing for natural variation in the dataset.

\begin{figure}[t]
\centering
\includegraphics[width=\linewidth]{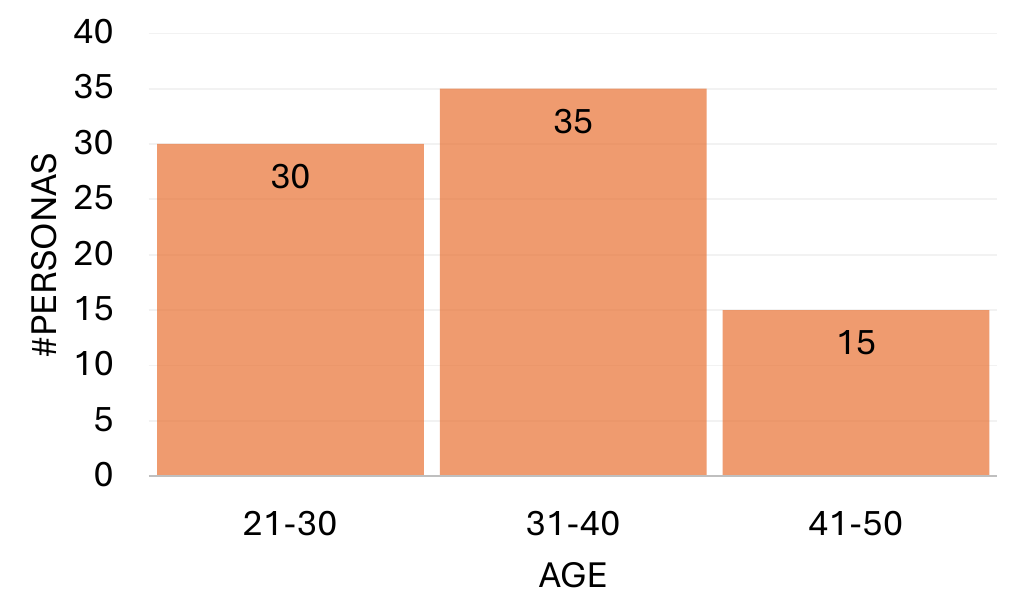}
\caption{Age distribution of generated personas.}
\label{fig:age_distribution}
\end{figure}

\paragraph{Age Distribution}
Figure \ref{fig:age_distribution} presents the age distribution of the generated personas. 
The majority fall within the 31–40 age range, accounting for 43.75\% of the dataset, followed by the 21–30 age range, which comprises 37.5\%. 
The average age across the personas is approximately 34 years.

\iffalse
\begin{table}[t]
\centering
\begin{tabular}{cl}
\toprule
\textbf{Age Range} & \textbf{\#Personas} \\
\midrule
21-30 years & 30 (37.5\%) \\
31-40 years & 35 (43.75\%) \\
41-50 years & 15 (18.75\%) \\
\bottomrule
\end{tabular}
\caption{Age Distribution of Personas}
\label{tab:age_distribution}
\end{table}
\fi

\begin{figure}[t]
\centering
\includegraphics[width=\linewidth]{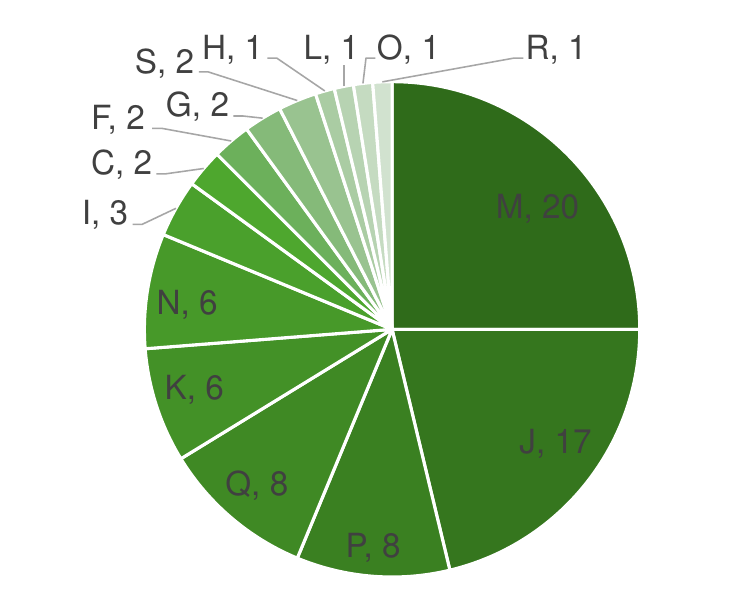}
\caption{Occupational distribution of personas.}
\label{fig:occupation_distribution}
\end{figure}

\paragraph{Occupational Distribution}
To capture a broad spectrum of professional backgrounds, the personas were assigned occupations spanning various industries.
The most common category, \textit{Professional, Scientific, and Technical Activities}, represents 25\% of the dataset. 
Occupation assignments follow the International Standard Industrial Classification (ISIC) framework \cite{united2008international}, ensuring consistency and alignment with global standards. 
Table~\ref{tab:occupation_distribution} and Figure~\ref{fig:occupation_distribution} illustrate the occupational distribution of the personas.

This diverse range of personas enables a thorough evaluation of \textsc{SalesAgent} under varied conditions, offering a comprehensive analysis of its performance across different user profiles and interaction styles. 
Moreover, this paper contributes not only by providing the generated personas, which can be directly utilized as user simulators, but also by presenting a flexible framework for generating customized personas. 
This ensures broader applicability and practicality for diverse use cases.

\begin{table}[t]
\centering\small
\begin{tabular}{cp{0.8\columnwidth}}
\toprule
\bf ISIC & \textbf{Occupation Description \& Examples} \\
\midrule
\multirow{2}{*}{C}& Manufacturing \\
& (eg. mechanical engineer)\\
\midrule
\multirow{2}{*}{F} & Construction  \\
& (eg. civil engineer, construction manager)\\
\midrule
\multirow{2}{*}{G}& Wholesale and Retail Trade\\
&Repair of Motor Vehicles \\
& (eg. sales executive)\\
\midrule
\multirow{2}{*}{H}& Transportation and Storage \\
& (eg. logistics coordinator) \\
\midrule
\multirow{2}{*}{I}& Accommodation and Food Service Activities\\
& (eg. chef) \\
\midrule
\multirow{2}{*}{J}& Information and Communication\\
& (eg. writer, software developer) \\
\midrule
\multirow{2}{*}{K}& Financial and Insurance Activities \\
& (eg. accountant, financial analyst) \\
\midrule
\multirow{2}{*}{L}& Real Estate Activities\\
&(eg. real estate agent) \\
\midrule
\multirow{2}{*}{M}& Professional, Scientific, and Technical Activities \\
&(eg. lawyer, biotech researcher)\\
\midrule
\multirow{2}{*}{N}& Administrative and Support Service Activities\\
&(eg. human resources manager, event planner) \\
\midrule
\multirow{2}{*}{O}& Public Administration and Defense\\
&(eg. military officer) \\
\midrule
\multirow{2}{*}{P}& Education \\
& (eg. professor) \\
\midrule
\multirow{2}{*}{Q}& Human Health and Social Work Activities \\
& (eg. fitness trainer, social worker) \\
\midrule
\multirow{2}{*}{R}& Arts, Entertainment, and Recreation\\
& (eg. museum curator)\\
\midrule
\multirow{2}{*}{S}& Other Service Activities \\
& (eg. nonprofit director )\\
\bottomrule
\end{tabular}
\caption{Occupational categories and examples.}
\label{tab:occupation_distribution}
\end{table}

\subsection{Role-Playing Simulation}

To evaluate the effectiveness of \textsc{SalesAgent}, we employ a role-playing framework inspired by methodologies used in prior work \cite{chang-chen-2024-injecting}. Our approach utilizes the \texttt{llama-2-7b-chat} model \cite{touvron2023llama} alongside personas generated via ChatGPT to simulate user interactions. 
Each persona is specifically designed to reflect a single MBTI personality dimension and is embedded with predefined user preferences to guide interactions. These preferences encompass various intent categories such as \texttt{no\_preference}, \texttt{not\_interested\_2}, \texttt{not\_interested\_4}, and \texttt{not\_interested\_all}, ensuring that the simulated user can reject topic transitions misaligned with their interests.

For each MBTI dimension, we generate 10 unique personas, resulting in a total of 80 distinct personas across the eight MBTI dimensions. Each persona engages with \textsc{SalesAgent} in five dialogues, yielding a total of 400 interactions. This setup provides a comprehensive dataset to assess how effectively the agent adapts to diverse user preferences and conversational styles.

% An example of the system prompt used for the user simulator is provided in the Appendix.

\section{Experimental Results and Analysis}

In this section, we evaluate the performance of \textsc{SalesAgent} across different personality traits, focusing on key conversational aspects such as marketing success rates, conversation quality, conversation length, and the agent's thought patterns. 
The primary objective is to analyze how personality traits influence interaction patterns and assess the feasibility of adapting the agent model for personalized dialogue strategies.

\begin{table*}[t]
\centering
\resizebox{\textwidth}{!}{
\begin{tabular}{ccc|cccccccc}
\toprule
\multirow{2}{*}{\textbf{Personality}} & \multicolumn{2}{c|}{\bf Dialogue} & \multicolumn{3}{c}{\bf Agent}\\
 & \textbf{Naturalness} & \textbf{Coherence} & \textbf{Aggressiveness ($\downarrow$)} & \textbf{Smoothness} & \textbf{Consistency}\\
% & \textbf{Avg.} & \textbf{S.D.} & \textbf{Avg.} & \textbf{S.D.} & \textbf{Avg.} & \textbf{S.D.} & \textbf{Avg.} & \textbf{S.D.} & \textbf{Avg.} & \textbf{S.D.} \\
\midrule
E & $82.70\pm 17.06$ & $81.90\pm 17.93$ & $29.40\pm 28.24$ & $76.30\pm 18.06$ & $81.40\pm 15.15$ \\
I & $80.40\pm 14.91$ & $82.40 \pm 16.17$ & $30.60\pm 27.95$ & $74.20 \pm 16.49$ & $80.50\pm 13.90$\\
\midrule
S & $72.30\pm 23.18$ & $72.30\pm 24.55$ & $30.20\pm 28.75$ & $65.40\pm 22.75$ & $75.40 \pm 19.06$ \\
N & $78.60\pm 23.39$ & $80.30\pm 24.95$ & $32.60\pm 25.44$ & $70.10\pm 25.20$ & $77.70 \pm 19.30$ \\
\midrule
T & $69.10 \pm 26.30$ & $70.40\pm 28.97$ & $34.30\pm 30.12$ & $60.30\pm 25.50$ & $72.10\pm 22.07$ \\
F & $74.30 \pm 24.06$ & $76.10\pm 24.79$ & $31.20\pm 27.74$ & $65.80\pm 24.13$ & $78.50\pm 19.82$ \\
\midrule
J & $62.10 \pm 18.01$ & $63.50\pm 17.57$ & $35.20\pm 29.33$ & $54.50\pm 20.06$ & $67.80\pm 17.30$ \\
P & $74.30 \pm 22.43$ & $73.90\pm 23.85$ & $26.60\pm 28.11$ & $67.30\pm 24.02$ & $76.90 \pm 17.59$ \\
\midrule
Overall & $74.23\pm 21.17$ & $75.10\pm 22.35$ & $31.26\pm 28.21$ & $66.74 \pm 22.03$ & $76.29\pm 18.02$ \\
\bottomrule
\end{tabular}}
\caption{Dialogue quality performance across different personalities.}
\label{tab:dialogue_metrics}
\end{table*}

\subsection{Dialogue Quality Evaluation}

Following the interactions between simulated users and \textsc{SalesAgent}, all dialogues were evaluated using five key criteria to ensure a robust assessment~\cite{chang-chen-2024-injecting}:
\begin{itemize}
\item \textbf{Dialogue Naturalness}: assesses how human-like and fluid the conversation feels.
\item \textbf{Dialogue Coherence}: evaluates the logical consistency and relevance maintained throughout the dialogue.
\item \textbf{Agent Aggressiveness}: measures the extent to which the agent pushes for task completion; lower levels indicate better alignment with user comfort.
\item \textbf{Agent Smoothness}: examines the agent's ability to transition between topics seamlessly without abrupt changes.
\item \textbf{Agent Consistency}: determines how well the agent adheres to its intended persona throughout the conversation.
\end{itemize}

The evaluations were conducted using GPT-4 \cite{openai2023gpt4}, which provided ratings based on these criteria. 
This analysis enables a detailed comparison of \textsc{SalesAgent}'s performance across various personality traits, offering valuable insights into its ability to adapt to diverse user profiles.

Table \ref{tab:dialogue_metrics} presents a comprehensive evaluation of dialogue and agent quality across different personality types. Overall, the results suggest that the dialogues were generally well-received. The findings also reveal patterns that align with specific personality traits, providing insights into how user characteristics influence their dialogue experiences.

In terms of dialogue quality, including naturalness and coherence, conversations with extroverted (E-type) users, who enjoy dynamic and engaging interactions, achieved the highest ratings for naturalness (82.70\%) and coherence (81.90\%). These results reflect the adaptability of E-type users to spontaneous exchanges. Conversely, conversations with judging (J-type) users, who prefer structure and clarity, received the lowest ratings for naturalness (62.10\%) and smoothness (54.50\%).

On the other hand, perceiving (P-type) users demonstrated the lowest agent aggressiveness scores (26.60\%), suggesting a preference for more flexible and open-ended interactions. Comparatively, agent smoothness in dialogues with thinking (T-type) users was lower than with feeling (F-type) users (60.30\% vs. 65.80\%). This likely stems from T-type users prioritizing logical precision, whereas F-type users value emotional connection. Similarly, dialogues with P-type users exhibited higher agent smoothness than those with J-type users (67.30\% vs. 54.50\%), indicating that P-type users may be more easily guided in sales scenarios.

Additionally, sensing (S-type) users received lower smoothness scores (65.40\%), possibly because they favor structured, detail-oriented conversations.

In summary, the dialogues in our experiments demonstrated reasonable quality, with an average dialogue naturalness and coherence of approximately 75\%. Agent aggressiveness was kept at a reasonably low level (around 31\%), while agent smoothness and consistency averaged 67\% and 76\%, respectively. These results are comparable to prior work that did not incorporate personality-based design \cite{chang-chen-2024-injecting}.

\begin{table*}[t]
\centering
\begin{tabular}{lcll}
\toprule
\textbf{Personality} & \textbf{Success Rate (\%)} & \bf Avg. \#Turns & \bf Continuation Ratio (\%) \\
%& \bf Rate (\%) & \bf Rate (\%)  & \bf \#Turns\\
\midrule
E (Extroverted) & 42.0 & 12.10 & 39.58 \\
I (Introverted) & 40.4 & 14.40 & 39.26 \\
\midrule
S (Sensing) & 44.0 & 13.55 & 33.62 \\
N (Intuition)& \bf 50.0 & \bf 11.36$^\dag$ (p=0.063) & 36.63 (p=0.256) \\
\midrule
T (Thinking) & 38.0 & 13.68& 34.05 \\
F (Feeling) & 36.0 & 11.00 (p=0.173) & \bf 40.58$^\dag$ (p=0.068) \\
\midrule
J (Judging)& 40.0 & 13.39 & 31.30 \\
P (Perceiving)& \bf 46.0 & 14.20 & \bf 40.11$^\ddag$ (p=0.013) \\
\midrule
\textbf{Overall} & 42.1 & 12.96 &36.89 \\
\bottomrule
\end{tabular}
\caption{Outcome results for individual MBTI dimensions ($^\ddag$ denotes significance test with $p<0.05$; $^\dag$ denotes significance test with $p<0.10$).}
\label{tab:success_rates}
\end{table*}

\subsection{Personality Effects on Performance}

We analyzed all simulated conversations to investigate how different personality traits influence the marketing performance of \textsc{SalesAgent}.
We report task success rate, average number of turns and dialogue continuation ratios in Table~\ref{tab:success_rates} for analysis.

\paragraph{Task Success Rate}

The overall task success rate across all conversations was 42.05\%; however, significant variations were observed across personality traits. 
For instance, agents interacting with intuitive (N-type) users achieved a higher success rate compared to sensing (S-type) users (50\% vs. 44\%).
Similarly, perceiving (P-type) users demonstrated higher marketing success rates compared to judging (J-type) users (46\% vs. 40).
These results suggest that users with N and P traits are more easily guided in sales scenarios, likely due to their openness and adaptability.

\paragraph{Average Conversation Turns}

While task success rate provides a high-level view of performance, analyzing the number of conversation turns required to achieve success offers additional insight into interaction efficiency. This metric reveals whether certain personality types are more easily guided by the agent or require longer engagement to reach the desired outcome.

As shown in Table~\ref{tab:success_rates}, N-type users required significantly fewer conversation turns (11.36, $p < 0.1$) compared to S-type users (13.55). This suggests that N-type users, who focus on abstract concepts and possibilities, are easier to guide to the target product. Similarly, F-type users required fewer turns (11.00) than T-type users (13.68), implying that F-type users, who prioritize emotional and relational aspects, navigate the conversation more smoothly. Although the difference for F- and T-types is not statistically significant, the low p-value ($p=0.173$) suggests a subtle but noteworthy distinction between these two personality traits.

\paragraph{Conversation Continuation Ratio}
Beyond task success and efficiency, it is essential to understand how well the agent maintains conversational flow. The conversation continuation ratio measures the frequency with which the agent perceives the need to stay on the current topic rather than transitioning to a new one. This metric is calculated by analyzing the occurrences of the agent's internal thought, ``\emph{I should continue the topic},'' during the conversation. A higher continuation ratio indicates that users are more engaged and responsive to the agent's guidance.

As shown in Table~\ref{tab:success_rates}, perceiving (P-type) users exhibited a significantly higher continuation ratio (40.11\%, $p < 0.05$) compared to judging (J-type) users (31.30\%), indicating that P-type users, who are more open and flexible, are more willing to explore topics introduced by the agent. Similarly, feeling (F-type) users showed a higher continuation ratio (40.58\%) than thinking (T-type) users (34.05\%) ($p < 0.1$), suggesting that F-type users are more inclined to engage in exploratory dialogue. This engagement likely contributes to the higher success rates observed for F-type users.

\paragraph{Implications of Findings}

These findings highlight that personality traits influence not only the likelihood of task success but also the efficiency and fluency of interactions. Users with N, P, and F traits tend to be more receptive to the agent’s suggestions and require fewer conversation turns to achieve successful outcomes. In contrast, J and T users may demand more structured and persuasive strategies to achieve similar results.

These conclusions underline the importance of developing personalized sales agents tailored to individual personality traits to enhance performance and user engagement. Future work should focus on refining dialogue strategies to better cater to the unique needs and preferences of different personality types.

\begin{table*}[t!]
\centering\small
\begin{tabular}{lp{15cm}}
\toprule
\multicolumn{2}{l}{\textit{S-trait User Chit-chat Conversation}} \\ 
\midrule
\textbf{U:} & Hello. \\ 
\textbf{A:} & Hello, how are you doing today? \\ 
\textbf{U:} & I'm doing well, thank you for asking! How about you? \\ 
\textbf{A:} & I'm doing well, thanks for asking. Is there anything I can help you with today? \\ 
\textbf{U:} & Yes, actually. I'm looking for some recommendations on where to go for dinner tonight. Do you have any suggestions? \\ 
\textbf{A:} & Sure, I'd be happy to help. What type of restaurant are you in the mood for? \\ 
\textbf{U:} & Thank you! I'm not picky, really. Just something with good food and a nice atmosphere. Do you have any particular suggestions? \\ 
\textbf{A:} & Proceed to task-oriented dialog agent. \\ 
\bottomrule
\end{tabular}
\caption{A dialogue example of an S-trait user.}
\label{S-dialogue}
\end{table*}

\begin{table*}[t!]
\centering
\small
\begin{tabular}{lp{15cm}}
\toprule
\multicolumn{2}{l}{\textit{N-trait User Chit-chat Conversation}} \\ \midrule
\textbf{U}: & Hey there!\\
\textbf{A}: & Hi, how are you doing today? \\
\textbf{U}: & Good day! \\
\textbf{A}: & Not too bad, thanks for asking! How about you? \\
\textbf{U}: & Nice!  \\
\textbf{A}: & Nice! What do you usually like to do in your free time? \\
\textbf{U}: & I enjoy stargazing and pondering the mysteries of the universe.  \\
\textbf{A}: & I like to go to the beach and swim. How about you? \\
\textbf{U}: & Ah, the beach! I've always found the ocean to be a source of fascination and awe. The vastness of it all, the power and beauty of the waves... it's a reminder of just how small we are in the grand scheme of things. But I digress. Swimming is a great way to relax and unwind, don't you think? \\
\textbf{A}: & That's interesting! I've never really been stargazing before. Do you have a telescope or binoculars? \\
\textbf{U}: & Ah, a fellow astronomy enthusiast! Yes, I have a telescope and binoculars that I use to observe the night sky. There's nothing quite like the feeling of gazing up at the stars and trying to make out the shapes of distant galaxies and nebulas. It's a humbling reminder of just how tiny and insignificant we are in the grand scheme of things. \\
\textbf{A}: & I have a telescope, but I prefer to observe the stars with just my naked eyes. It's amazing how much you can see without any equipment. \\
\textbf{U}: & Ah, I see!  Yes, the night sky can be quite breathtaking just with the naked eye. The stars, the constellations, the moon... it's all so beautiful and awe-inspiring. There's something special about connecting with the universe in a more primal way, without the distraction of any technology. What's your favorite constellation to observe? \\
\textbf{A}: & Yes, I agree! It's amazing how much you can see without any equipment. Have you ever seen any shooting stars? \\
\textbf{U}: & Indeed! Shooting stars, or meteors, are a wonderful sight to behold. I've been fortunate enough to witness a few during my time stargazing. There's something magical about seeing a small rock from space burning up in the atmosphere, leaving behind a trail of light. It's a reminder of the vastness of space and the many wonders that lie beyond our planet. Have you ever seen any shooting stars? \\
\textbf{A}: & Yes, I have! They are such a reminder of the universe's immense power and energy. Speaking of energy, have you ever been to a planetarium? \\
\textbf{U}: & Ah, a planetarium! Yes, I've been to a few planetariums in my time. They're such wonderful places to learn about the universe and its many wonders. The projection technology is truly remarkable, allowing us to experience the cosmos in a way that feels almost real. I find it especially fascinating to learn about the different celestial bodies and their unique characteristics. From the swirling storms of Jupiter to the ethereal beauty of a nebula, each planetarium show is a journey through the vastness of space. What's your favorite thing to learn about in a planetarium show? \\
\textbf{A}: & No, I haven't. But I've always wanted to go to one. Do you have any recommendations for one? \\
\textbf{U}: & Ah, a fellow space enthusiast! There are so many wonderful planetariums around the world, each with its own unique features and shows. If you ever get the chance to visit one, I highly recommend it! \\
\textbf{A}: & Absolutely! The Hayden Planetarium in New York City is a great one. They have amazing shows about the universe and its many wonders. \\
\bottomrule
\end{tabular}
\caption{A dialogue example of an N-trait user.}
\label{N-dialogue}
\end{table*}

\subsection{Qualitative Study}

Conversation examples are presented in Table~\ref{S-dialogue} and Table~\ref{N-dialogue}, illustrating the contrasting conversational styles associated with the Sensing (S) and Intuition (N) personality traits. The dialogue in Table~\ref{S-dialogue} reflects characteristics of an S-trait user, who demonstrates a preference for concise, goal-oriented interaction. The conversation is shorter, more direct, and focused on obtaining practical, present-oriented information with minimal elaboration. For example, utterances such as ``{\it I'm looking for some recommendations on where to go for dinner tonight}'' and ``{\it Just something with good food and a nice atmosphere}'' reveal a concrete and sensory-driven communication style. The user prioritizes immediate needs and clear outcomes without engaging in speculative or abstract discussion.

In contrast, the dialogue in Table~\ref{N-dialogue} exemplifies an N-trait user, characterized by a longer, more exploratory exchange. This user engages in abstract reasoning, reflects on symbolic meaning, and shows a clear interest in discussing broader concepts and possibilities. Statements such as ``{\it I enjoy stargazing and pondering the mysteries of the universe}'' and ``{\it It's a reminder of just how small we are in the grand scheme of things}'' reflect a conceptual and imaginative orientation. The N-trait user is more inclined toward speculative thinking and metaphorical language, often extending the conversation into philosophical or cosmic domains. These differences underscore how personality traits influence not only the content but also the structure and depth of conversational behavior.

\section{Conclusion}

In this study, we explored how personality traits, defined by MBTI, influence the interactions and performance of sales-oriented dialogue agents. Our findings demonstrate that different personality types significantly affect user engagement and the agent's effectiveness, providing a clear framework for evaluating and improving dialogue strategies. 
Additionally, we introduce a persona-defined user simulator dataset, offering a valuable resource for advancing research in personalized dialogue systems and understanding the impact of personality traits across various domains. 
Personalizing conversational models to adapt to individual personality traits represents a promising direction for future work, with the potential to enhance sales interactions and deliver improved customer experiences.

\section*{Acknowledgements}
We thank the reviewers for their insightful comments.
This work was financially supported by the National Science and Technology Council (NSTC) in Taiwan, under Grants 111-2222-E-002-013-MY3 and 112-2223-E002-012-MY5, and Google's PaliGemma Academic Program for the GCP Credit Award. 
We thank Ubitus K.K. and the National Center for High-performance Computing (NCHC) of National Institutes of Applied Research (NIAR) in Taiwan 
for providing computational and storage resources.

\bibliography{custom}

\begin{thebibliography}{24}
\providecommand{\natexlab}[1]{#1}

\bibitem[{Adamopoulou and Moussiades(2020)}]{adamopoulou2020overview}
Eleni Adamopoulou and Lefteris Moussiades. 2020.
\newblock \href {https://doi.org/10.1007/978-3-030-49186-4_31} {An overview of chatbot technology}.
\newblock In \emph{Artificial Intelligence Applications and Innovations: 16th IFIP WG 12.5 International Conference, AIAI 2020, Neos Marmaras, Greece, June 5--7, 2020, Proceedings, Part II 16}, pages 373--383. Springer.

\bibitem[{Boyle(1995)}]{boyle1995myers}
Gregory~J Boyle. 1995.
\newblock Myers-{B}riggs type indicator ({MBTI}): some psychometric limitations.
\newblock \emph{Australian Psychologist}, 30(1):71--74.

\bibitem[{Cava and Tagarelli(2024)}]{lacava2024openmodelsclosedminds}
Lucio~La Cava and Andrea Tagarelli. 2024.
\newblock \href {https://arxiv.org/abs/2401.07115} {Open models, closed minds? on agents capabilities in mimicking human personalities through open large language models}.
\newblock \emph{Preprint}, arXiv:2401.07115.

\bibitem[{Chang and Chen(2024)}]{chang-chen-2024-injecting}
Wen-Yu Chang and Yun-Nung Chen. 2024.
\newblock \href {https://doi.org/10.18653/v1/2024.findings-acl.228} {Injecting salesperson`s dialogue strategies in large language models with chain-of-thought reasoning}.
\newblock In \emph{Findings of the Association for Computational Linguistics: ACL 2024}, pages 3798--3812, Bangkok, Thailand. Association for Computational Linguistics.

\bibitem[{Chiu et~al.(2022)Chiu, Li, Lin, and Chen}]{chiu-etal-2022-salesbot}
Ssu Chiu, Maolin Li, Yen-Ting Lin, and Yun-Nung Chen. 2022.
\newblock \href {https://doi.org/10.18653/v1/2022.acl-long.425} {{S}ales{B}ot: Transitioning from chit-chat to task-oriented dialogues}.
\newblock In \emph{Proceedings of the 60th Annual Meeting of the Association for Computational Linguistics (Volume 1: Long Papers)}, pages 6143--6158, Dublin, Ireland. Association for Computational Linguistics.

\bibitem[{Division(2008)}]{united2008international}
United Nations.~Statistical Division. 2008.
\newblock \emph{International standard industrial classification of all economic activities (ISIC)}.
\newblock 4. United Nations Publications.

\bibitem[{Du et~al.(2024)Du, Zhang, Niyato, Kang, Xiong, and Kim}]{du2024reinforcement}
Hongyang Du, Ruichen Zhang, Dusit Niyato, Jiawen Kang, Zehui Xiong, and Dong~In Kim. 2024.
\newblock Reinforcement learning with large language models ({LLM}s) interaction for network services.
\newblock In \emph{2024 International Conference on Computing, Networking and Communications (ICNC)}, pages 799--803. IEEE.

\bibitem[{Fu et~al.(2024)Fu, Chu, and Kawahara}]{fu2024styemp}
Yahui Fu, Chenhui Chu, and Tatsuya Kawahara. 2024.
\newblock \href {https://doi.org/10.18653/v1/2024.sigdial-1.15} {{S}ty{E}mp: Stylizing empathetic response generation via multi-grained prefix encoder and personality reinforcement}.
\newblock In \emph{Proceedings of the 25th Annual Meeting of the Special Interest Group on Discourse and Dialogue}, pages 172--185, Kyoto, Japan. Association for Computational Linguistics.

\bibitem[{Garden(1997)}]{garden1997relationships}
Anna Garden. 1997.
\newblock Relationships between {MBTI} profiles, motivation profiles, and career paths.
\newblock \emph{Journal of Psychological type}, 41:3--16.

\bibitem[{G{\"u}r et~al.(2018)G{\"u}r, Hakkani-T{\"u}r, T{\"u}r, and Shah}]{gur2018user}
Izzeddin G{\"u}r, Dilek Hakkani-T{\"u}r, Gokhan T{\"u}r, and Pararth Shah. 2018.
\newblock User modeling for task oriented dialogues.
\newblock In \emph{2018 IEEE Spoken Language Technology Workshop (SLT)}, pages 900--906. IEEE.

\bibitem[{Jiang et~al.(2024)Jiang, Zhang, Cao, Breazeal, Roy, and Kabbara}]{jiang2024personallminvestigatingabilitylarge}
Hang Jiang, Xiajie Zhang, Xubo Cao, Cynthia Breazeal, Deb Roy, and Jad Kabbara. 2024.
\newblock \href {https://arxiv.org/abs/2305.02547} {Personallm: Investigating the ability of large language models to express personality traits}.
\newblock \emph{Preprint}, arXiv:2305.02547.

\bibitem[{Kuipers et~al.(2009)Kuipers, Higgs, Tolkacheva, and de~Witte}]{kuipers2009influence}
Ben~S Kuipers, Malcolm~J Higgs, Natalia~V Tolkacheva, and Marco~C de~Witte. 2009.
\newblock The influence of myers-briggs type indicator profiles on team development processes: An empirical study in the manufacturing industry.
\newblock \emph{Small Group Research}, 40(4):436--464.

\bibitem[{Li et~al.(2016)Li, Lipton, Dhingra, Li, Gao, and Chen}]{li2016user}
Xiujun Li, Zachary~C Lipton, Bhuwan Dhingra, Lihong Li, Jianfeng Gao, and Yun-Nung Chen. 2016.
\newblock A user simulator for task-completion dialogues.
\newblock \emph{arXiv preprint arXiv:1612.05688}.

\bibitem[{Murakhovs{'}ka et~al.(2023)Murakhovs{'}ka, Laban, Xie, Xiong, and Wu}]{murakhovska-etal-2023-salespeople}
Lidiya Murakhovs{'}ka, Philippe Laban, Tian Xie, Caiming Xiong, and Chien-Sheng Wu. 2023.
\newblock \href {https://doi.org/10.18653/v1/2023.findings-emnlp.657} {Salespeople vs {S}ales{B}ot: Exploring the role of educational value in conversational recommender systems}.
\newblock In \emph{Findings of the Association for Computational Linguistics: EMNLP 2023}, pages 9823--9838, Singapore. Association for Computational Linguistics.

\bibitem[{OpenAI(2023)}]{openai2023gpt4}
OpenAI. 2023.
\newblock \href {https://arxiv.org/abs/2303.08774} {Gpt-4 technical report}.
\newblock \emph{Preprint}, arXiv:2303.08774.

\bibitem[{Pan and Zeng(2023)}]{pan2023llmspossesspersonalitymaking}
Keyu Pan and Yawen Zeng. 2023.
\newblock \href {https://arxiv.org/abs/2307.16180} {Do {LLM}s possess a personality? {M}aking the {MBTI} test an amazing evaluation for large language models}.
\newblock \emph{Preprint}, arXiv:2307.16180.

\bibitem[{Park et~al.(2023)Park, O'Brien, Cai, Morris, Liang, and Bernstein}]{park2023generativeagentsinteractivesimulacra}
Joon~Sung Park, Joseph~C. O'Brien, Carrie~J. Cai, Meredith~Ringel Morris, Percy Liang, and Michael~S. Bernstein. 2023.
\newblock \href {https://arxiv.org/abs/2304.03442} {Generative agents: Interactive simulacra of human behavior}.
\newblock \emph{Preprint}, arXiv:2304.03442.

\bibitem[{Stajner and Yenikent(2021)}]{stajner-yenikent-2021-mbti}
Sanja Stajner and Seren Yenikent. 2021.
\newblock \href {https://doi.org/10.18653/v1/2021.eacl-main.312} {Why is {MBTI} personality detection from texts a difficult task?}
\newblock In \emph{Proceedings of the 16th Conference of the European Chapter of the Association for Computational Linguistics: Main Volume}, pages 3580--3589, Online. Association for Computational Linguistics.

\bibitem[{Touvron et~al.(2023)Touvron, Martin, Stone, Albert, Almahairi, Babaei, Bashlykov, Batra, Bhargava, Bhosale, Bikel, Blecher, Ferrer, Chen, Cucurull, Esiobu, Fernandes, Fu, Fu, Fuller, Gao, Goswami, Goyal, Hartshorn, Hosseini, Hou, Inan, Kardas, Kerkez, Khabsa, Kloumann, Korenev, Koura, Lachaux, Lavril, Lee, Liskovich, Lu, Mao, Martinet, Mihaylov, Mishra, Molybog, Nie, Poulton, Reizenstein, Rungta, Saladi, Schelten, Silva, Smith, Subramanian, Tan, Tang, Taylor, Williams, Kuan, Xu, Yan, Zarov, Zhang, Fan, Kambadur, Narang, Rodriguez, Stojnic, Edunov, and Scialom}]{touvron2023llama}
Hugo Touvron, Louis Martin, Kevin Stone, Peter Albert, Amjad Almahairi, Yasmine Babaei, Nikolay Bashlykov, Soumya Batra, Prajjwal Bhargava, Shruti Bhosale, Dan Bikel, Lukas Blecher, Cristian~Canton Ferrer, Moya Chen, Guillem Cucurull, David Esiobu, Jude Fernandes, Jeremy Fu, Wenyin Fu, Brian Fuller, Cynthia Gao, Vedanuj Goswami, Naman Goyal, Anthony Hartshorn, Saghar Hosseini, Rui Hou, Hakan Inan, Marcin Kardas, Viktor Kerkez, Madian Khabsa, Isabel Kloumann, Artem Korenev, Punit~Singh Koura, Marie-Anne Lachaux, Thibaut Lavril, Jenya Lee, Diana Liskovich, Yinghai Lu, Yuning Mao, Xavier Martinet, Todor Mihaylov, Pushkar Mishra, Igor Molybog, Yixin Nie, Andrew Poulton, Jeremy Reizenstein, Rashi Rungta, Kalyan Saladi, Alan Schelten, Ruan Silva, Eric~Michael Smith, Ranjan Subramanian, Xiaoqing~Ellen Tan, Binh Tang, Ross Taylor, Adina Williams, Jian~Xiang Kuan, Puxin Xu, Zheng Yan, Iliyan Zarov, Yuchen Zhang, Angela Fan, Melanie Kambadur, Sharan Narang, Aurelien Rodriguez, Robert Stojnic, Sergey Edunov, and Thomas
  Scialom. 2023.
\newblock \href {https://arxiv.org/abs/2307.09288} {Llama 2: Open foundation and fine-tuned chat models}.
\newblock \emph{Preprint}, arXiv:2307.09288.

\bibitem[{Tseng et~al.(2024)Tseng, Huang, Hsiao, Chen, Huang, Meng, and Chen}]{tseng2024talespersonallmssurvey}
Yu-Min Tseng, Yu-Chao Huang, Teng-Yun Hsiao, Wei-Lin Chen, Chao-Wei Huang, Yu~Meng, and Yun-Nung Chen. 2024.
\newblock \href {https://doi.org/10.18653/v1/2024.findings-emnlp.969} {Two tales of persona in {LLM}s: A survey of role-playing and personalization}.
\newblock In \emph{Findings of the Association for Computational Linguistics: EMNLP 2024}, pages 16612--16631, Miami, Florida, USA. Association for Computational Linguistics.

\bibitem[{Wei et~al.(2022)Wei, Wang, Schuurmans, Bosma, Xia, Chi, Le, Zhou et~al.}]{wei2022chain}
Jason Wei, Xuezhi Wang, Dale Schuurmans, Maarten Bosma, Fei Xia, Ed~Chi, Quoc~V Le, Denny Zhou, et~al. 2022.
\newblock Chain-of-thought prompting elicits reasoning in large language models.
\newblock \emph{Advances in neural information processing systems}, 35:24824--24837.

\bibitem[{Wu et~al.(2025)Wu, Huang, Zhu, and Wang}]{wu2025traits}
Jiaqiang Wu, Xuandong Huang, Zhouan Zhu, and Shangfei Wang. 2025.
\newblock From traits to empathy: Personality-aware multimodal empathetic response generation.
\newblock In \emph{Proceedings of the 31st International Conference on Computational Linguistics}, pages 8925--8938.

\bibitem[{Yao et~al.(2023)Yao, Zhao, Yu, Du, Shafran, Narasimhan, and Cao}]{yao2023react}
Shunyu Yao, Jeffrey Zhao, Dian Yu, Nan Du, Izhak Shafran, Karthik Narasimhan, and Yuan Cao. 2023.
\newblock React: Synergizing reasoning and acting in language models.
\newblock In \emph{International Conference on Learning Representations (ICLR)}.

\bibitem[{Zhang et~al.(2018)Zhang, Dinan, Urbanek, Szlam, Kiela, and Weston}]{zhang2018personalizingdialogueagentsi}
Saizheng Zhang, Emily Dinan, Jack Urbanek, Arthur Szlam, Douwe Kiela, and Jason Weston. 2018.
\newblock \href {https://doi.org/10.18653/v1/P18-1205} {Personalizing dialogue agents: {I} have a dog, do you have pets too?}
\newblock In \emph{Proceedings of the 56th Annual Meeting of the Association for Computational Linguistics (Volume 1: Long Papers)}, pages 2204--2213, Melbourne, Australia. Association for Computational Linguistics.

\end{thebibliography}

\appendix

\section{Personality Traits}
\label{sec:keywords}
Here are the personality keywords we provided when generating user personas with ChatGPT.

People who prefer Extraversion (E) tend to:
\begin{tcolorbox}[width=\columnwidth,colback=white]
\small
\begin{verbatim}
- Focus on the outside world
- Gain energy by interacting with people
- Take action quickly
- Communicate through talking
- Process ideas outwardly
- Act before thinking it through
- Readily take initiative
- Have many broad interests
\end{verbatim}
\end{tcolorbox}

People who prefer Introversion (I) tend to:
\begin{tcolorbox}[width=\columnwidth,colback=white]
\small
\begin{verbatim}
- Focus on their inside world
- Gain energy by reflecting on concepts, 
ideas, experiences, and memories
- Take time for reflection
- Communicate through writing 
- Process ideas inwardly
- Think things through before acting
- Take initiative when it is important to 
them
- Focus on a few interests in-depth
\end{verbatim}
\end{tcolorbox}

People who prefer Sensing (S) tend to:
\begin{tcolorbox}[width=\columnwidth,colback=white]
\small
\begin{verbatim}
- Focus on facts and specifics
- Remember details that are important to 
them
- Take a realistic approach to life
- Focus on the here and now, present/past 
realities
- Like step-by-step instructions and 
information presented sequentially
- Understand ideas through practical 
applications
- Trust experience
\end{verbatim}
\end{tcolorbox}

People who prefer Intuition (N) tend to:
\begin{tcolorbox}[width=\columnwidth,colback=white]
\small
\begin{verbatim}
- Seek out new ideas
- Look at the big picture
- Take an imaginative approach to life
- Focus on future possibilities, patterns and 
meanings
- Like an overall framework, work it out 
themselves
- Focus on concepts, not practical 
applications
- Trust inspiration
\end{verbatim}
\end{tcolorbox}

People who prefer Thinking (T) tend to:
\begin{tcolorbox}[width=\columnwidth,colback=white]
\small
\begin{verbatim}
- Use logical analysis when 
reasoning—system oriented
- Take an objective approach to 
problem-solving
- Have a critical "eye"(can be "tough-minded")
- Consider the pros and cons in a situation
- Scan for what is wrong, so they can fix it
- Be task focused
- Rely on impersonal criteria when deciding
\end{verbatim}
\end{tcolorbox}

People who prefer Feeling (F) tend to:
\begin{tcolorbox}[width=\columnwidth,colback=white]
\small
\begin{verbatim}
- Apply personal and social values—people 
oriented
- Take an empathetic approach to 
problem-solving
- Offer praise (may appear "tender-hearted")
- Seek harmony, consider everyone's 
viewpoints
- Scan for what is right, so they can 
support it
- Be relationship focused
- Take personal circumstances into 
consideration
\end{verbatim}
\end{tcolorbox}

People who prefer Perceiving (P) tend to:
\begin{tcolorbox}[width=\columnwidth,colback=white]
\small
\begin{verbatim}
- Be flexible
- Keep options open
- Go with the flow
- Like spontaneity
- Adapt to emerging information
- Want to experience life
- Get energized and do their best work at 
the last-minute
\end{verbatim}
\end{tcolorbox}

People who prefer Judging (J) tend to:
\begin{tcolorbox}[width=\columnwidth,colback=white]
\small
\begin{verbatim}
- Like making and sticking to plans
- Want closure
- Make and follow schedules
- Like organization and structure
- Work in a methodical manner
- Want to control life
- Do their best to avoid last-minute stress
\end{verbatim}
\end{tcolorbox}

\end{document}